\documentclass[12pt]{article}

\usepackage{amsmath}
\usepackage{graphicx}
\usepackage{booktabs}
\usepackage{natbib}
\usepackage{hyperref}
\usepackage{geometry}
\usepackage{mathptmx} 
\usepackage{multirow}
\usepackage{array}
\usepackage{placeins}

\hypersetup{
  colorlinks=true,
  linkcolor=blue,
  citecolor=blue,
  urlcolor=blue
}

\title{Trajectory Dynamics in Language Model Hidden States Predict Human Processing Costs Beyond Surprisal}

\author{Elan Barenholtz\\
Machine Perception \& Cognitive Robotics Laboratory\\
Department of Psychology / Center for Complex Systems\\
Florida Atlantic University\\
\texttt{elan.barenholtz@fau.edu}}

\date{}

\begin{document}

\maketitle

\begin{abstract}
Human language comprehension unfolds sequentially: each word is processed in the context of those that came before, and the interpretation builds incrementally over time. Surprisal, the negative log probability of a word given its context, has been the dominant predictor of incremental processing cost. But surprisal reduces rich sequential representations to a single scalar at each word, discarding information about the direction in which the interpretation has been evolving. Dynamical-systems approaches suggest that the trajectory of the evolving interpretive state, not just its position at each moment, should shape processing, and language itself may have short-horizon continuity, since speakers plan utterances a few words at a time. We introduce trajectory extrapolation error: at each word, we fit a linear trajectory to the preceding hidden states of a transformer language model and measure deviation from the extrapolated path. On the Natural Stories corpus, this measure is nearly orthogonal to surprisal ($r = .044$) and independently predicts self-paced reading times. The effect is present in garden-path sentences and replicates across GPT-2 variants (Small, Medium, Large) and across architectures with different positional encoding schemes (GPT-2 vs.\ Pythia/RoPE). A displacement control shows the effect is not reducible to representational change magnitude: displacement and extrapolation error predict in opposite directions. These findings reveal two dissociable components of processing cost: word-level prediction error (surprisal) and sensitivity to local continuity in the evolving interpretive representation (trajectory extrapolation error).

\end{abstract}

\newpage

\section{Introduction}

Human language comprehension unfolds sequentially. Words arrive one at a time, and the reader or listener must build an interpretation incrementally, each word updating and extending the representation constructed from those that came before. Understanding this incremental process is a central problem in psycholinguistics, and the word-by-word variation in reading times has long served as its primary empirical signature. A core regularity in this signature is that words predictable in context are processed more quickly than words that are not. The dominant computational framework for explaining this regularity is surprisal theory \citep{hale2001,levy2008}, which holds that the cost of processing a word is proportional to its negative log probability given the preceding context. The framework follows naturally from information-theoretic considerations: if the comprehender maintains a probability distribution over upcoming words, then encountering a low-probability word requires a larger update to that distribution, and this update cost is what the reader experiences as difficulty. Surprisal has been remarkably successful as a predictor of reading times across self-paced reading \citep{smith2013}, eye-tracking \citep{demberg2008}, and neural measures \citep{frank2015}.

Computing surprisal requires an underlying predictive model. Early work used n-gram models and probabilistic grammars \citep{hale2001,demberg2008,roark2009}, which provided useful but limited estimates of word predictability. Transformer-based language models have since become the standard tool \citep{goodkind2018,wilcox2020,schrimpf2021}, largely because their surprisal estimates are better predictors of human reading times than those from earlier models, and because they can condition on arbitrarily long contexts rather than fixed windows. They are also a natural fit because they are themselves sequential processors: trained to predict the next token given the preceding sequence, they build rich internal representations that update incrementally as each word is encountered. Because these models are trained on human-produced text, the statistical structure they capture --- sensitivity to context, coherence, discourse --- is the same structure latent in that production, which is what makes their surprisal a useful index of the predictive regularities to which human comprehenders are sensitive.

But sequential processing may involve more than step-by-step prediction. An alternative theoretical tradition treats comprehension not as a series of independent predictions but as a dynamical process in which the interpretive state evolves through a continuous representational space \citep{tabor1999,spivey2007,cho2017}. Under this account, the trajectory of the evolving state, not just its position at each moment, carries information that shapes processing. This is especially likely for language, because human language production itself is a sequential, locally planned process. Speakers and writers plan a few words ahead, execute that plan, and then re-plan \citep{levelt1989,ferreira2002}, creating text with local momentum: stretches over which the context evolves in a coherent direction before shifting. If this momentum is a real property of natural language, and if comprehenders are sensitive to it, then the direction in which the interpretation has been evolving, not just the probability of the next word, should matter for processing.

Testing whether trajectory dynamics shape human processing requires a way to measure trajectory structure in an evolving interpretive representation. Modern transformer language models like GPT-2 \citep{radford2019} provide one. Their hidden states encode a rich, incrementally updated representation of the context processed so far, and because the model is trained on human-produced text, these representations come to reflect not only the predictive structure that surprisal captures, but also other latent regularities of production --- including the short-horizon continuity of how each word's representation extends from those of its predecessors. Here, we introduce a simple measure that we call trajectory extrapolation error. At each word position, we fit a linear trajectory to the hidden states of the preceding $k$ words (typically $k = 3$), extrapolate one step forward, and measure the Euclidean distance between this predicted position and the actual hidden state. The measure captures the degree to which the current word deviates from where the representation was heading: high extrapolation error means the representation was moving in one direction and the current word forced it somewhere else; low error means the current word continued the established drift.

The logic of comparing trajectory extrapolation error against surprisal is asymmetric. Surprisal is expected to predict processing cost under either tradition, since improbable words tend to disrupt trajectories; finding that surprisal matters does not distinguish between the frameworks. Finding that trajectory extrapolation error adds independent explanatory power beyond surprisal would, by contrast, be informative: it would suggest that the dynamical character of comprehension contributes to processing cost in a way that word-level prediction error does not capture. Two words with identical conditional probability receive identical surprisal even if one continues an established interpretive trajectory and the other forces a sharp representational turn, and any cost difference between such words is evidence for directional dynamics in comprehension. This asymmetry is not a limitation of the language models from which surprisal is derived: their hidden states already carry the trajectory information, but surprisal as an output measure collapses it to a scalar. The question this paper takes up is whether the dynamics that surprisal discards are psychologically real.

There are empirical reasons to think they do. Human processing unfolds under strong constraints of recency and local influence \citep{gibson1998,lewis2005}, meaning that recent words dominate the current interpretive state. Under such constraints, the trajectory of the interpretation over the last few words is the comprehender's best available signal about where things are heading. Sensitivity to this trajectory, tracking the recent drift of the interpretation rather than treating each word as an independent event, would constitute an efficient processing strategy that exploits the local momentum of natural language. Garden-path sentences provide an intuitive illustration: in ``The horse raced past the barn fell,'' each word after ``raced'' reinforces the main-verb interpretation, building momentum in one direction, and the cost at ``fell'' reflects not just the word's unexpectedness but the reversal of accumulated interpretive direction. But if trajectory sensitivity is a general feature of human processing, the phenomenon should not be limited to garden paths. Any word that forces the interpretation off its recent trajectory should incur a cost, even in ordinary text, and this cost should be measurable independently of surprisal.

A puzzling empirical finding reinforces this possibility. \citet{oh2023b} demonstrated that as language models grow larger and achieve lower perplexity, their surprisal estimates become \textit{worse} predictors of human reading times, a phenomenon known as the surprisal scaling paradox. Predictive power peaks at modest model sizes and then diverges from human behavior. This is consistent with a trajectory-based account: as models improve at full-context prediction, their surprisal values increasingly reflect a processing strategy that is less and less like the recency-dominated, trajectory-sensitive processing that humans may actually be doing.
The present study assesses whether human reading times reflect such directional sensitivity beyond what surprisal captures. We compute trajectory extrapolation error from the hidden states of transformer language models and ask whether it contributes to reading-time prediction in two complementary datasets: the Classic Garden Path subset of the SAP Benchmark \citep{huang2024}, where the disambiguation point provides a controlled, theoretically motivated locus of trajectory disruption, and the Natural Stories corpus \citep{futrell2018}, where the same measure is evaluated over thousands of word positions in naturalistic text. The diagnostic test is whether trajectory extrapolation error contributes to reading-time prediction independently of surprisal. An affirmative result would indicate that directional dynamics constitutes a dissociable dimension of processing cost rather than a redundant summary of word-level predictability. Three additional analyses clarify what the measure captures. A displacement control assesses whether the contribution of extrapolation error reduces to the magnitude of representational change at each word. A direction-preservation analysis on the Natural Stories corpus characterizes the temporal scale of the underlying trajectory structure in the model. Finally, multi-model comparisons across GPT-2 Small, Medium, and Large, and a cross-architecture replication using Pythia (which uses Rotary Position Embeddings rather than the absolute positional embeddings of GPT-2), test whether any effect of trajectory structure generalizes across model scale and positional encoding scheme.

\section{Method}

\subsection{Materials}

\paragraph{Garden-path sentences.} To test the trajectory effect at a controlled, theoretically motivated locus of disruption, we used the Classic Garden Path subset of the SAP Benchmark \citep{huang2024}, a large-scale syntactic ambiguity processing dataset. The subset contains 24 items covering three structural types: main-verb/reduced-relative (MVRR; e.g., ``The horse raced past the barn fell''), NP/S direct-object/sentential-complement (e.g., ``The suspect showed the file deserved more attention''), and NP/Z transitive/intransitive (e.g., ``While the man hunted the deer ran into the woods''). Each item appeared in ambiguous and unambiguous conditions, where the unambiguous version included an overt syntactic marker (e.g., a relative pronoun) that prevents the garden-path misparse. Human reading-time data were collected via self-paced word-by-word reading from over 2,000 participants. Reaction times were filtered to exclude responses below 100 ms or above 5,000 ms. Analyses focused on the critical region: the disambiguating word and two subsequent spillover positions (i.e., the two words immediately following the disambiguation point, where processing difficulty typically persists as the reader completes reanalysis).

\paragraph{Natural Stories.} To evaluate the same measure across thousands of word positions in naturalistic text, we used the Natural Stories corpus \citep{futrell2018}, which consists of 10 naturalistic narratives totaling approximately 10,000 words. Self-paced reading times were collected from 181 participants, yielding 845,479 observations after filtering to 100--3,000 ms. We processed each story through GPT-2 in overlapping chunks to accommodate the model's 1,024-token context limit, computing hidden states and surprisal at every word position.

\subsection{Trajectory Extrapolation Error}

\begin{figure}[!ht]
\centering
\includegraphics[width=\textwidth]{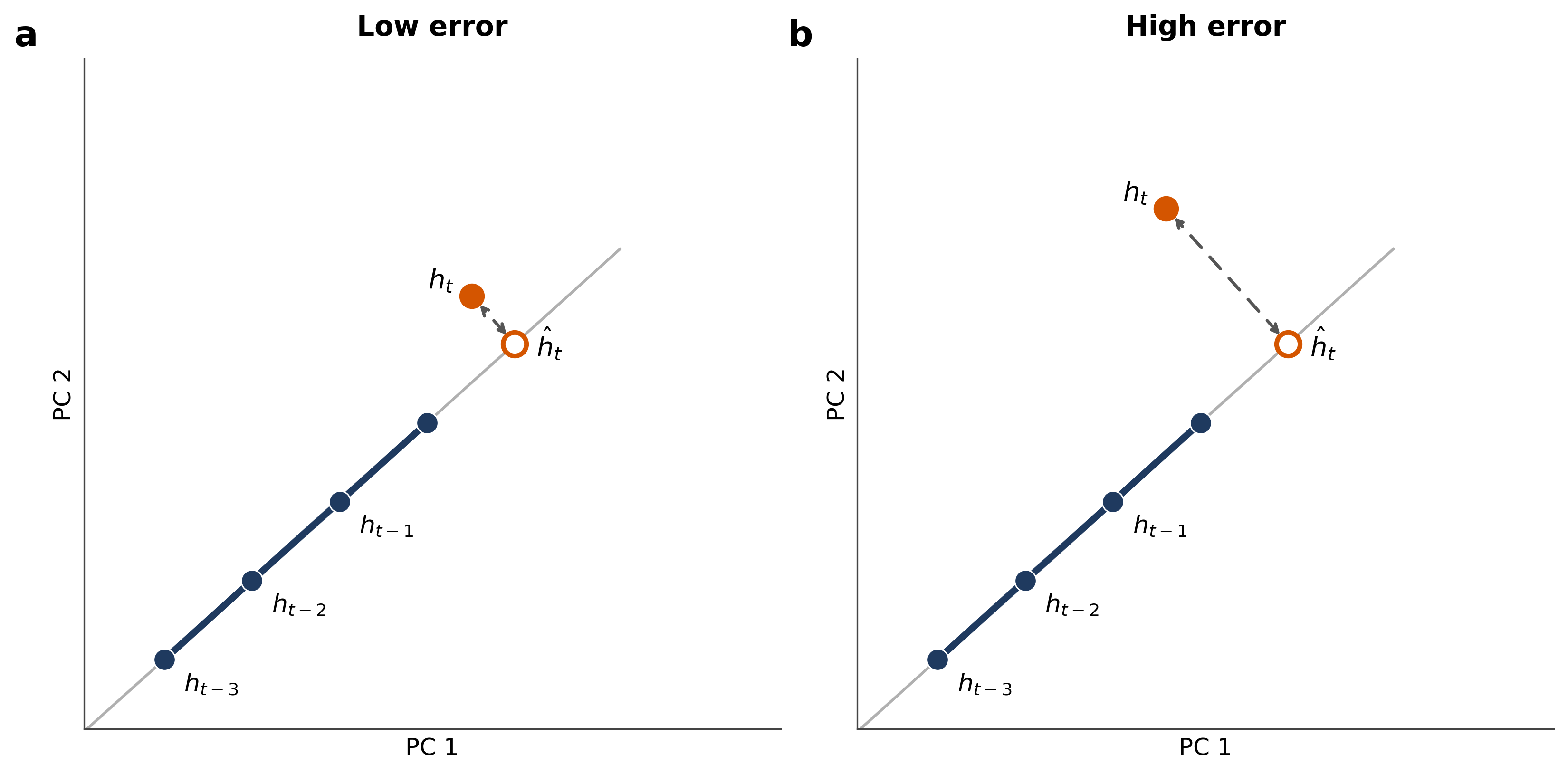}
\caption{Schematic illustration of trajectory extrapolation error. Each point $h_t$ represents the model's hidden-state vector at word position $t$, projected into a two-dimensional principal-component space for visualization. A linear trajectory is fit to the hidden states at the preceding $k$ positions (here $k = 3$: $h_{t-3}$ through $h_{t-1}$) and extrapolated one step forward to produce a predicted position (open circle, $\hat{h}_t$). Trajectory extrapolation error is the Euclidean distance between this predicted position and the actual hidden state $h_t$ (filled orange circle). (a) Low error: the current word's hidden state falls near the extrapolated position, continuing the established trajectory. (b) High error: the current word (e.g., a garden-path disambiguator) forces the hidden state far from the extrapolated position.}
\label{fig:schematic}
\end{figure}

\paragraph{Definition.} Let $h_t$ denote the hidden-state vector at word position $t$ in a given layer of a transformer language model. For a window of size $k$, we fit a linear trajectory to the hidden states at positions $t{-}k$ through $t{-}1$ by ordinary least squares. The extrapolated position at the next time step is the linear prediction at time $k$, and trajectory extrapolation error is defined as the Euclidean distance between the extrapolated and actual hidden states (see Figure~\ref{fig:schematic} for a schematic illustration). This quantity measures how far the representation landed from where it was heading, given the trajectory established by the preceding $k$ words.

\paragraph{Parameter selection.} We computed extrapolation error using GPT-2 (117M parameters; \citealp{radford2019}), which sits near the sweet spot for reading-time prediction identified by \citet{oh2023a}. We tested two layers: layer 6 (an intermediate layer) and the final layer (layer 12). We focus primarily on layer 6 because intermediate layers are known to capture syntactic representations more effectively than output layers in transformer models \citep{hewitt2019,jawahar2019}. We swept window sizes of 3, 5, and 7 words with both linear and quadratic polynomial fits (see Table~\ref{tab:gardenpath} for the full comparison). The 3-word linear fit at layer 6 consistently outperformed longer windows and higher-degree fits, consistent with the subsequent finding that trajectory structure in the hidden states is strictly local. For sentences with subword tokenization, we used the hidden state at the last subword token of each word.

\subsection{Direction Preservation Analysis}

To characterize the trajectory structure underlying extrapolation error, we computed a direction preservation measure on the Natural Stories corpus. At each word position, we measured the absolute cosine similarity between the fitted trajectory direction (from the preceding $k$ words) and the actual displacement vector at the current word and at 1, 2, and 3 steps ahead. This measures whether the direction of representational change at one position predicts the direction of change at subsequent positions. For random vectors in 768-dimensional space, the expected absolute cosine similarity is approximately 0.029, providing a chance baseline.

\subsection{Statistical Analyses}

We fit linear mixed-effects models predicting log-transformed reading times, with random intercepts for participants. All continuous predictors were $z$-scored prior to entry. The baseline model (M0) included word length, word position, previous log RT, and log word frequency (Zipf score; standard psycholinguistic control, e.g., \citealp{smith2013}) as lexical controls. We then tested a series of nested models: M1 added surprisal; M2 added trajectory extrapolation error. Model comparisons were evaluated using the Akaike Information Criterion (AIC), the Bayesian Information Criterion (BIC), and likelihood ratio tests. AIC estimates out-of-sample prediction error, with lower values indicating better fit; BIC applies a stronger penalty for model complexity and is more conservative. For the garden-path analysis, the critical region included the disambiguating word and two spillover positions (ROI codes 0, 1, and 2 in the SAP Benchmark coding scheme). For the Natural Stories analysis, we additionally included log word frequency as a control and tested the independence of surprisal and extrapolation error via correlation and a tercile-based dissociation analysis. As a further control, we tested whether extrapolation error reduces to simple one-step representational change by comparing it with embedding displacement ($\|h_t - h_{t-1}\|$) in a joint model including both measures alongside surprisal and lexical controls.

To test the robustness of the effect across model scale, we repeated the garden-path reading-time analysis using GPT-2 Medium (345M parameters, 24 layers) and GPT-2 Large (774M parameters, 36 layers), testing the proportionally equivalent mid-layer (50\% depth: layer 12 for Medium, layer 18 for Large) and the final layer with window sizes of 3 and 5.

As a validation, we compared extrapolation error at the disambiguating word between ambiguous and unambiguous conditions in the garden-path dataset. If the measure captures anything meaningful about garden-path processing, ambiguous sentences should show substantially higher extrapolation error at the disambiguation point than their unambiguous counterparts.

\section{Results}

\subsection{Garden-Path Sentences}

\paragraph{Validation.} Ambiguous sentences showed substantially higher extrapolation error at the disambiguation point than their unambiguous counterparts across all layer and window configurations, confirming that the measure is sensitive to the garden-path manipulation.

\paragraph{Reading-time prediction.} Adding surprisal to a control model that included log word frequency did not improve fit (Table~\ref{tab:gardenpath}, M1: $\Delta\text{AIC} = -1.9$, $\chi^2(1) = 0.14$, $p = .71$). Adding trajectory extrapolation error at layer 6 with a 3-word window did (Table~\ref{tab:gardenpath}, M2: $\Delta\text{AIC} = 10.7$, $\chi^2(1) = 12.7$, $p = 3.7 \times 10^{-4}$; $\beta = +0.005$). The improvement was robust across several configurations of layer and window size (Table~\ref{tab:gardenpath}, M3--M5).

\begin{table}[!htbp]
\centering
\caption{Model Comparison for Garden-Path Reading-Time Prediction}
\label{tab:gardenpath}
\begin{tabular}{llrrrrl}
\toprule
\textbf{Model} & \textbf{Predictors} & \textbf{AIC} & $\boldsymbol{\Delta}$\textbf{AIC} & $\boldsymbol{\Delta}$\textbf{BIC} & $\boldsymbol{\chi^2(1)}$ & $\boldsymbol{p}$ \\
\midrule
M0 & Controls (incl. log freq) & 94,124.7 & --- & --- & --- & --- \\
M1 & M0 + Surprisal & 94,126.6 & $-1.9$ & $-11.3$ & 0.14 & .71 \\
M2 & M1 + Extrap error (L6, $w = 3$) & 94,115.9 & 10.7 & 1.2 & 12.7 & $3.7 \times 10^{-4}$ \\
M3 & M1 + Extrap error (L12, $w = 5$) & 94,070.2 & 56.4 & 46.9 & 58.4 & $<10^{-10}$ \\
M4 & M1 + Extrap error (L6, $w = 5$) & 94,126.6 & 0.0 & $-9.4$ & 2.0 & .15 \\
M5 & M1 + Extrap error (L6, $w = 7$) & 94,095.3 & 31.4 & 21.9 & 33.4 & $7.7 \times 10^{-9}$ \\
\bottomrule
\end{tabular}

\smallskip
\footnotesize
\textit{Note.} $N = 95{,}173$ observations across 2,000+ participants. Controls = word length + word position + previous log RT + log word frequency. $\Delta$AIC and $\Delta$BIC are improvements over the preceding nested model (M1 for surprisal over M0; M2--M5 over M1). L6 = layer 6; L12 = final layer. BIC penalizes model complexity more heavily than AIC; the negative $\Delta$BIC for surprisal indicates that BIC does not justify its inclusion.
\end{table}

\subsection{Natural Stories}

\paragraph{Independence of measures.} The correlation between surprisal and extrapolation error at the word level was $r = .044$, indicating that the two measures are nearly orthogonal. Whatever trajectory extrapolation error captures, it is not a nonlinear transformation of surprisal.

\begin{figure}[t]
\centering
\includegraphics[width=0.8\textwidth]{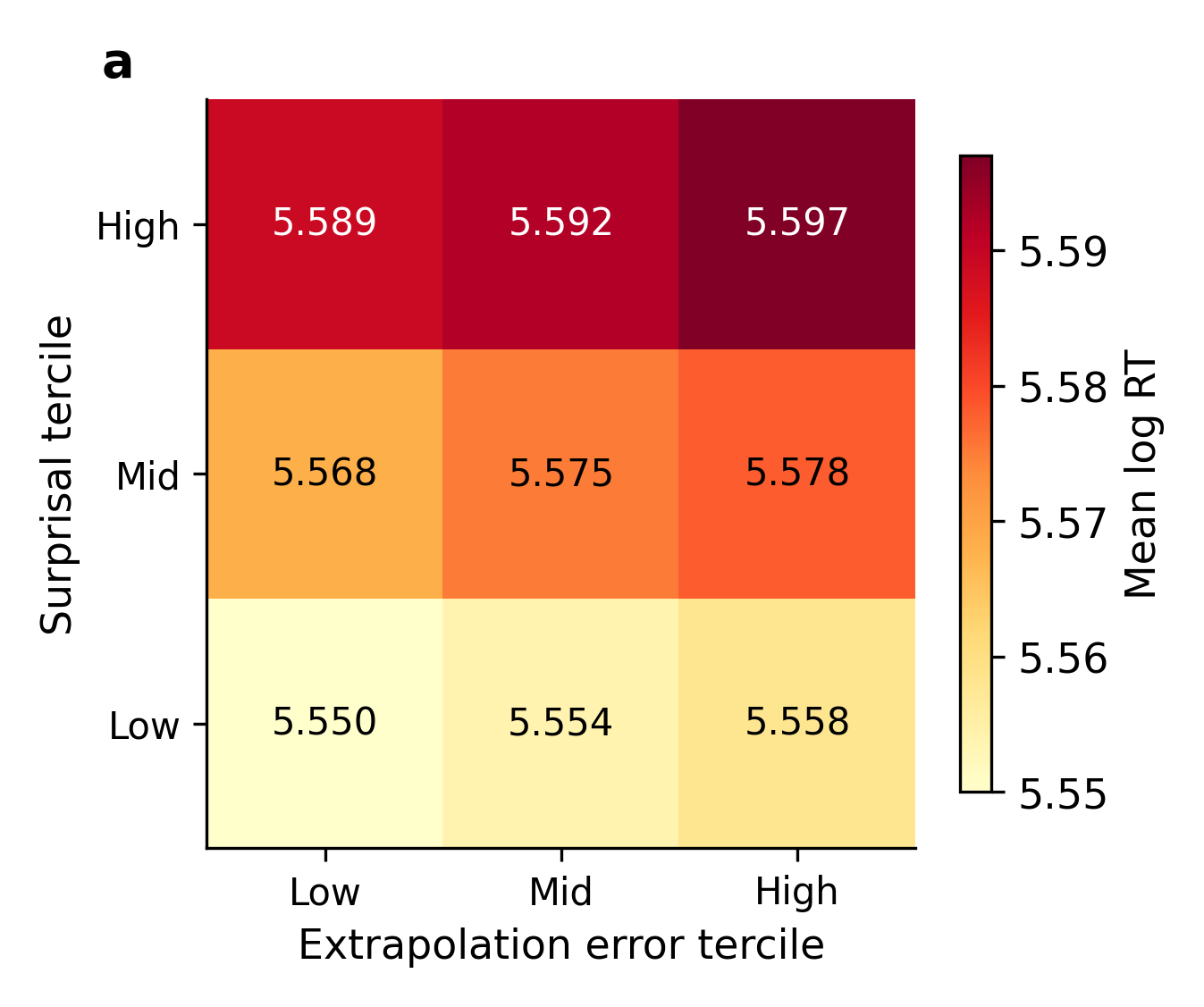}
\caption{Dissociation between surprisal and trajectory extrapolation error in Natural Stories. Heatmap shows mean log reading time in each cell of the $3 \times 3$ tercile matrix. The off-diagonal cells demonstrate that each measure captures unique variance: high surprisal / low extrapolation error (surprise without trajectory disruption) and low surprisal / high extrapolation error (trajectory disruption without surprise) both produce elevated reading times relative to the low/low baseline.}
\label{fig:dissociation}
\end{figure}

\paragraph{Dissociation analysis.} We split words into terciles on both surprisal and extrapolation error and examined mean log reading times in each cell of the resulting $3 \times 3$ matrix (Table~\ref{tab:dissociation}; Figure~\ref{fig:dissociation}). Words with high surprisal but low extrapolation error (surprising words that continue the trajectory) showed an increase of $+0.039$ in log RT over the low/low baseline ($t = 21.24$, $p < 10^{-100}$). Words with low surprisal but high extrapolation error (unsurprising words that disrupt the trajectory) showed an increase of $+0.008$ ($t = 4.67$, $p = 3.0 \times 10^{-6}$). Both effects are significant, confirming that each measure captures unique variance in reading times.

\paragraph{Regression.} A mixed-effects model including both $z$-scored surprisal and $z$-scored extrapolation error confirmed independent, additive effects. The additive model was preferred over a model including their interaction ($\Delta\text{AIC} = -2.0$ in favor of the simpler model), indicating that surprisal and extrapolation error operate through different mechanisms that combine without amplifying or dampening each other.

\begin{table}[!htbp]
\centering
\caption{Dissociation Matrix: Mean Log Reading Time by Surprisal $\times$ Extrapolation Error Tercile (Natural Stories)}
\label{tab:dissociation}
\begin{tabular}{lccc}
\toprule
 & \textbf{Low extrap error} & \textbf{Mid extrap error} & \textbf{High extrap error} \\
\midrule
\textbf{Low surprisal} & baseline & + & +0.008 \\
\textbf{Mid surprisal} & + & + & + \\
\textbf{High surprisal} & +0.039 & + & +0.047 \\
\bottomrule
\end{tabular}

\smallskip
\footnotesize
\textit{Note.} Off-diagonal cells represent the key dissociation: high surprisal / low extrapolation error reflects surprise without trajectory disruption; low surprisal / high extrapolation error reflects trajectory disruption without surprise. Both off-diagonal effects are independently significant (see text).
\end{table}

\subsection{Displacement Control}

To test whether extrapolation error reduced to simple representational change, we compared it with one-step displacement, $\|h_t - h_{t-1}\|$. The two measures were only weakly correlated ($r = .16$). Moreover, they made independent contributions to reading time when entered jointly with surprisal and lexical controls (Table~\ref{tab:displacement}: displacement beyond extrapolation error and surprisal: $\chi^2(1) = 8.30$, $p = .004$; extrapolation error beyond displacement and surprisal: $\chi^2(1) = 6.32$, $p = .012$). Critically, their effects had opposite signs (Table~\ref{tab:displacement}): displacement predicted faster reading ($\beta = -0.008$, $p = .004$), whereas extrapolation error predicted slower reading ($\beta = +0.001$, $p = .012$). Thus, processing cost is not associated with representational movement per se. Rather, large changes appear facilitative when they continue the local trajectory, while deviations from the extrapolated trajectory incur a cost. This rules out the simplest alternative interpretation, that extrapolation error merely indexes the magnitude of representational change, and supports the interpretation of extrapolation error as a measure of trajectory violation rather than simple representational disruption.

\begin{table}[!htbp]
\centering
\caption{Displacement Control: Joint Model of One-Step Displacement and Extrapolation Error on Natural Stories Reading Times}
\label{tab:displacement}
\begin{tabular}{lrrr}
\toprule
\textbf{Predictor} & $\boldsymbol{\beta}$ & $\boldsymbol{\chi^2(1)}$ & $\boldsymbol{p}$ \\
\midrule
Displacement ($\|h_t - h_{t-1}\|$) & $-0.008$ & 8.30 & .004 \\
Extrapolation error & $+0.001$ & 6.32 & .012 \\
\bottomrule
\end{tabular}

\smallskip
\footnotesize
\textit{Note.} Both predictors entered into a joint mixed-effects model alongside surprisal and lexical controls. Correlation between displacement and extrapolation error: $r = .16$. The opposite-sign coefficients confirm that extrapolation error captures trajectory violation rather than the magnitude of representational change at each word.
\end{table}

\subsection{Direction Preservation}

\begin{figure}[t]
\centering
\includegraphics[width=0.8\textwidth]{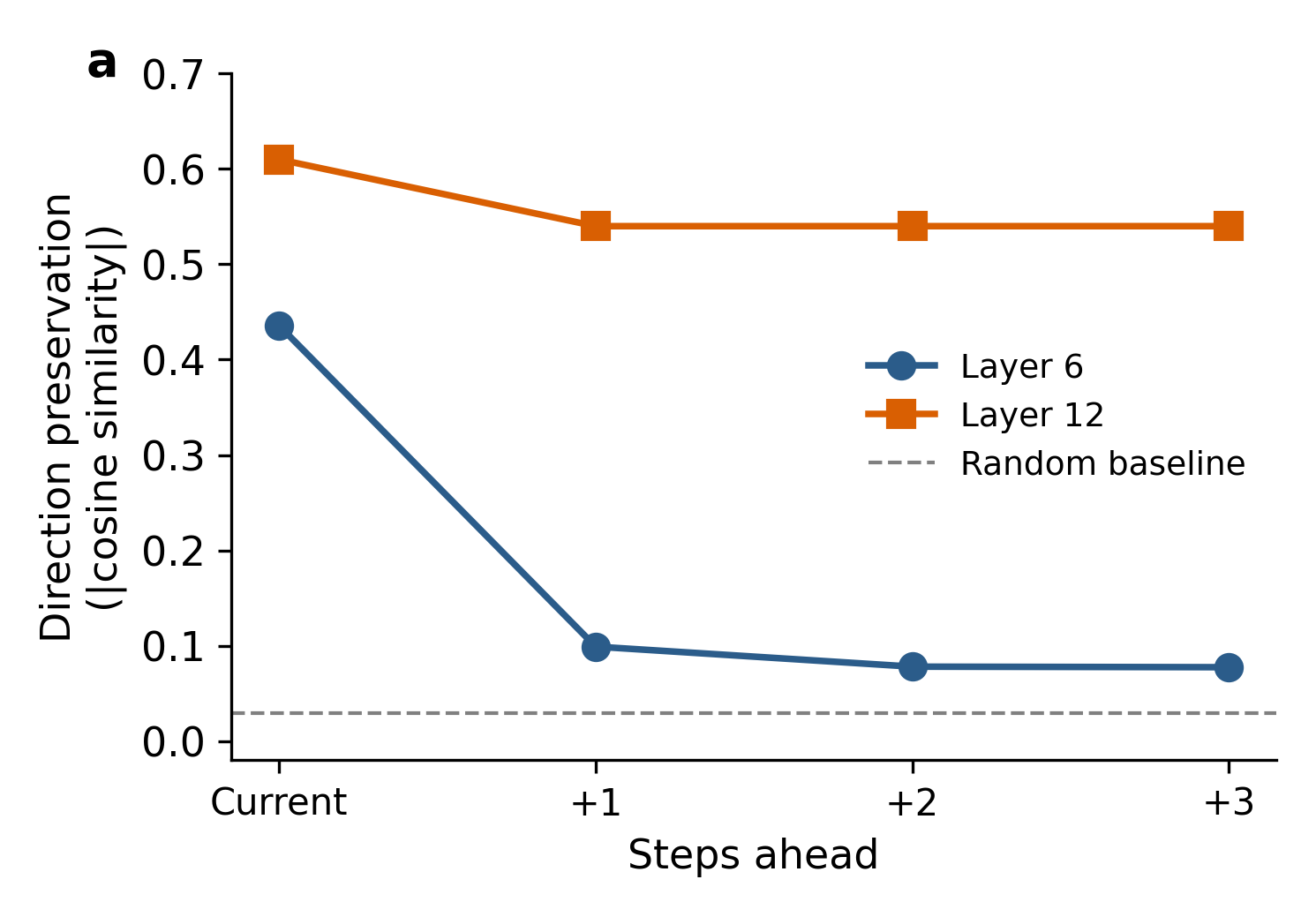}
\caption{Direction preservation decay by layer and steps ahead (Natural Stories). At layer 6, direction preservation drops from 0.44 at the current step to near-random levels (0.10) at one step ahead, indicating that the trajectory structure is strictly local. At the final layer (layer 12), direction preservation remains elevated (0.54) across all steps ahead, indicating persistent directional structure. Dashed horizontal line indicates the random baseline for 768-dimensional vectors (0.029).}
\label{fig:direction}
\end{figure}

The results reveal a clear dissociation across layers (Table~\ref{tab:direction}; Figure~\ref{fig:direction}). At layer 6, direction preservation at the current word position is 0.44, well above the random baseline of 0.029. However, preservation drops to 0.10 at one step ahead and remains at approximately 0.08 for two and three steps ahead. Direction effectively dies after a single word. At the final layer, direction preservation at the current step is 0.61 and remains at 0.54 across one, two, and three steps ahead. The final layer maintains persistent directional structure that the intermediate layer does not.

This finding clarifies what extrapolation error captures at layer 6. Because direction dies after one step, the linear fit over 3 words does not capture a persistent trajectory; rather, it captures local positional continuity --- the smooth displacement of hidden states due to the incremental nature of contextual updating. Higher-surprisal words showed lower direction preservation ($r = -0.10$ at layer 6), confirming that surprising words disrupt the trajectory at both positional and directional levels.

\begin{table}[!htbp]
\centering
\caption{Direction Preservation in Natural Stories by Layer and Steps Ahead}
\label{tab:direction}
\begin{tabular}{lccccc}
\toprule
\textbf{Layer} & \textbf{Current step} & \textbf{+1 step} & \textbf{+2 steps} & \textbf{+3 steps} & \textbf{Random baseline} \\
\midrule
Layer 6 ($w = 3$) & 0.44 & 0.10 & 0.08 & 0.08 & 0.029 \\
Layer 12 ($w = 3$) & 0.61 & 0.54 & 0.54 & 0.54 & 0.029 \\
Layer 6 ($w = 5$) & 0.39 & 0.09 & 0.08 & 0.07 & 0.029 \\
Layer 12 ($w = 5$) & 0.58 & 0.53 & 0.53 & 0.53 & 0.029 \\
\bottomrule
\end{tabular}

\smallskip
\footnotesize
\textit{Note.} Values represent mean absolute cosine similarity between the fitted trajectory direction and the displacement vector at each word position. Random baseline is the expected absolute cosine similarity between random vectors in 768-dimensional space.
\end{table}

\subsection{Robustness Across Model Scale}

To test whether the trajectory extrapolation effect is specific to GPT-2 Small (117M parameters) or reflects a more general property of learned representations, we repeated the garden-path reading-time analysis using GPT-2 Medium (345M parameters, 24 layers) and GPT-2 Large (774M parameters, 36 layers). For each model, we tested the proportionally equivalent mid-layer (layer 12 for Medium, layer 18 for Large) and the final layer, with window sizes of 3 and 5.

The contribution of trajectory extrapolation error replicated across GPT-2 variants (Table~\ref{tab:multimodel}): $\Delta\text{AIC} = 10.7$ for Small ($p = 3.7 \times 10^{-4}$), 1.6 for Medium (marginal, $p = .058$), and 13.1 for Large ($p = 1.0 \times 10^{-4}$). The effect is present at all three model sizes and is not specific to any one of them.

\begin{table}[!htbp]
\centering
\caption{Robustness Across Model Scale: Best Configuration per Model (Mid-Layer, $w = 3$)}
\label{tab:multimodel}
\resizebox{\textwidth}{!}{%
\begin{tabular}{lcrrrl}
\toprule
\textbf{Model} & \textbf{Params} & \textbf{$\Delta$AIC surp / ctrl} & \textbf{$\Delta$AIC extrap / surp} & \textbf{$\Delta$AIC surp / extrap} & $\boldsymbol{\chi^2(1)}$ \\
\midrule
GPT-2 Small (L6) & 117M & $-1.9$ & 10.7 & $-1.2$ & 12.7 ($p = 3.7 \times 10^{-4}$) \\
GPT-2 Medium (L12) & 345M & 6.5 & 1.6 & 7.4 & 3.6 ($p = .058$) \\
GPT-2 Large (L18) & 774M & 18.7 & 13.1 & 21.9 & 15.1 ($p = 1.0 \times 10^{-4}$) \\
\bottomrule
\end{tabular}%
}

\smallskip
\footnotesize
\textit{Note.} All models include log word frequency, word length, word position, and previous log RT as controls. $\Delta$AIC surp / ctrl = improvement from adding surprisal to the baseline control model. $\Delta$AIC extrap / surp = improvement from adding trajectory extrapolation error to the model already containing surprisal. $\Delta$AIC surp / extrap = improvement from adding surprisal to the model already containing extrapolation error (the reverse direction); negative values indicate the addition is not justified. All models tested on the same 24 garden-path items with reading times from 2,000+ participants. Mid-layer refers to the proportionally equivalent layer at the 50\% depth point of each model.
\end{table}

\subsection{Cross-Architecture Replication}

The preceding analyses all use GPT-2 variants, which share the same architecture and learned absolute positional embeddings. To test whether the trajectory extrapolation effect depends on this specific positional encoding scheme, we replicated the Natural Stories analysis using Pythia models \citep{biderman2023}, which use Rotary Position Embeddings (RoPE), a fundamentally different approach that encodes position through rotation of the hidden-state vectors rather than through additive position embeddings. We tested Pythia-160M (comparable to GPT-2 Small; mid-layer 6 of 12) and Pythia-410M (comparable to GPT-2 Medium; mid-layer 12 of 24).

The core finding replicated across both architectures (Table~\ref{tab:crossarch}). Trajectory extrapolation error predicted reading times beyond surprisal in both Pythia-160M ($\Delta\text{AIC} = +67.0$, $\chi^2(1) = 69.01$, $p < 10^{-16}$) and Pythia-410M ($\Delta\text{AIC} = +13.3$, $\chi^2(1) = 15.34$, $p < 10^{-4}$). The near-orthogonality of surprisal and extrapolation error was essentially identical to GPT-2: $r = .046$ for Pythia-160M and $r = .047$ for Pythia-410M, compared to $r = .044$ for GPT-2. Direction preservation showed the same rapid decay at the mid-layer, dropping from 0.42 to near-random levels within a single step for both Pythia models, confirming that the strictly local trajectory structure is not an artifact of GPT-2's coordinate system.

\begin{table}[!htbp]
\centering
\caption{Cross-Architecture Replication: Pythia (RoPE) vs.\ GPT-2 (Absolute Position Embeddings)}
\label{tab:crossarch}
\begin{tabular}{lccrcc}
\toprule
\textbf{Model} & \textbf{Pos.\ enc.} & $\boldsymbol{r}$\textbf{(surp, extrap)} & \textbf{$\Delta$AIC} & \textbf{Dir.\ pres.\ +0} & \textbf{Dir.\ pres.\ +1} \\
\midrule
GPT-2 Small (117M) & Absolute & .044 & +2.5 & 0.44 & 0.10 \\
Pythia-160M & RoPE & .046 & +67.0 & 0.42 & 0.13 \\
Pythia-410M & RoPE & .047 & +13.3 & 0.42 & 0.11 \\
\bottomrule
\end{tabular}

\smallskip
\footnotesize
\textit{Note.} $\Delta$AIC extrap / surprisal = improvement from adding trajectory extrapolation error to the model already containing surprisal and lexical controls. GPT-2 Small result is from the full Natural Stories dataset (180 participants); Pythia results are from a subsample of 100 participants. Direction preservation values are mean absolute cosine similarity between the fitted trajectory direction and the displacement vector at the current (+0) and next (+1) word positions.
\end{table}

The GPT-2 displacement control (opposite-sign dissociation between displacement and extrapolation error) did not replicate with Pythia, where both measures predicted in the same direction. This is expected: the specific relationship between one-step displacement and multi-step trajectory deviation depends on the representational geometry, which differs across positional encoding schemes. The theoretically motivated finding, that trajectory extrapolation error captures processing-relevant information beyond surprisal, is architecture-general.

\FloatBarrier

\section{Discussion}

\subsection{Two Dimensions of Sequential Processing}

Surprisal-based accounts have been remarkably productive at predicting incremental processing costs, but they treat comprehension as a sequence of evaluations against a probability distribution and discard the temporal structure of how the interpretation has been evolving. The results presented here suggest that human readers are sensitive to that dynamical structure as well. Specifically, our results show that human language processing times are impacted by two dissociable properties of the incoming signal: the probability of each word given its context (surprisal) as well as the degree to which each word deviates from the short-horizon trajectory of the evolving representation (trajectory extrapolation error). Critically, the effect of trajectory extrapolation replicates across multiple datasets as well as diverse model architectures with fundamentally different positional encoding schemes (GPT-2's absolute embeddings and Pythia's Rotary Position Embeddings), confirming that it reflects genuine properties of the hidden-state dynamics rather than an artifact of any particular model's coordinate system.

Interestingly, surprisal and trajectory extrapolation error are nearly orthogonal ($r = .044$ in Natural Stories), making independent and additive contributions to reading times. If both measures are derived from the same model processing the same text, why are they so independent? The answer is that they capture errors at different levels of the sequential process. Surprisal reflects the model's token-level prediction error: how unlikely was this specific word, given the full context? Trajectory extrapolation error reflects the representational reorientation cost: how much did the hidden state change direction, given where it had been heading? A word can be probable but trajectory-breaking (it was expected but redirects the interpretation), or improbable but trajectory-continuing (it was unlikely but keeps the representation moving in the same direction). The two dimensions are logically independent, and the data confirm that they are empirically independent as well.

The displacement control analysis in GPT-2 rules out the simplest alternative interpretation of the trajectory-extrapolation error finding: that it merely indexes the magnitude of representational change in the embedding space. In GPT-2, displacement and extrapolation error predict in opposite directions: large representational changes facilitate processing when they continue the trajectory, while deviations from the trajectory impede it. This confirms that the measure captures something specifically about local trajectory continuity rather than change magnitude. This opposite-sign pattern did not replicate in Pythia, where both measures predict in the same direction, suggesting that the specific relationship between displacement and trajectory deviation depends on the representational geometry of the model. However, the core finding, that trajectory extrapolation error contributes to reading-time prediction beyond surprisal, is robust across architectures.

The trajectory effect in Natural Stories is small in absolute terms (the coefficient is modest and the variance explained is a fraction of what surprisal contributes), but it is theoretically diagnostic: it is consistent, independent of surprisal, and directionally specific in a way that the simplest alternatives do not predict. The effect is more pronounced at controlled loci of trajectory disruption: in the garden-path data, trajectory extrapolation error contributes to reading-time prediction beyond surprisal and lexical controls at the disambiguation point and spillover ($\Delta\text{AIC} \approx 10\text{--}13$ in GPT-2 Small and Large after frequency control). This is consistent with the interpretation that what the measure indexes is the cost of trajectory deviation: garden-path sentences epitomize this kind of disruption, with the ambiguous region building an interpretive trajectory in one direction and the disambiguating word forcing a sharp reversal of that direction. At minimum, the data point to two dissociable components operating at different timescales: one that evaluates each word against a probability distribution, and another that is sensitive to the short-horizon trajectory structure of the interpretation as it evolves across words.

These results support a more dynamical process of language comprehension in which the trajectory of the evolving interpretive state, not just its position at each moment, shapes how each new word is processed. However, our results specifically support local representational continuity over a few-word window, not long-range directional momentum: the direction-preservation analysis confirms that directional structure dies within a single word at the predictive layer, so the trajectory the measure captures is strictly short-horizon. This pattern suggests that a single mechanism, whether framed as prediction error, information-theoretic surprise, or Bayesian updating, does not fully account for the costs of incremental comprehension.

\subsection{Trajectory as a Functional Property of Language}

Why might language comprehension track this short-term trajectory structure? The answer, we suggest, lies in both how language is produced and processed. Human language production is a sequential, locally planned process: speakers and writers plan a few words ahead, execute that plan, and then re-plan \citep{levelt1989,ferreira2002}. This creates text with local momentum: stretches over which the context evolves coherently in a particular direction before shifting. This momentum is not incidental. It is what maintains local coherence in the signal. The direction in which the interpretation has been moving carries information about the topic, the argument, the narrative arc, and this information would be lost if the representation were recomputed from scratch at every word. In this sense, trajectory is not merely a statistical regularity of text; it is a functional property of language that serves coherence. The 3-word window that dominates our results may correspond roughly to the planning horizon of human language production, and the trajectory structure at this timescale may reflect the sequential footprint of the production process itself.

Language models learn this short-horizon trajectory structure because it is present in their training data: human-produced text carries the local continuity of human production planning. The model's hidden states come to reflect this structure as a statistical byproduct of learning to predict the next word well. But the model does not use trajectory for its own processing; the direction-preservation analysis shows that at the predictive layer, each word is effectively a fresh computation, with near-zero directional persistence from one step to the next. The model recomputes from full context at every position. The trajectory is in the representation, not in the processing strategy.

From the standpoint of comprehension, language processing also proceeds under strong recency constraints and limited working memory \citep{gibson1998,lewis2005}, so re-deriving the full interpretation from the entire preceding context at every word is not a tractable strategy. Some compressed summary of recent context is computationally necessary, and the local trajectory of the evolving interpretation is a natural candidate: it summarizes where the interpretation has been going over the last few words. A word that continues this local trajectory is cheap to integrate; a word that forces a deviation is expensive. The empirical question we addressed is whether human reading times are sensitive to this kind of local-continuity signal beyond what surprisal already captures.

Garden-path sentences are where the local continuity fails most dramatically. The local trajectory established by the ambiguous region points in the wrong direction, and the disambiguating word reveals that what looked like a coherent local trajectory was misleading. Thus, we observed the strongest effect for these sentence types. But the phenomenon is not limited to garden paths. Any word that forces the representation off its recent trajectory incurs a cost, even in ordinary text, as demonstrated by the Natural Stories analysis.

This framing connects to lossy-context surprisal \citep{futrell2020}, which proposes that comprehenders work with a lossy representation of context rather than a perfect one. Trajectory extrapolation is a specific, extreme form of lossy context, perhaps the lossiest summary imaginable: a linear fit to three points in high-dimensional space. It also connects to the surprisal scaling paradox \citep{oh2023b}: as language models grow larger, their surprisal estimates reflect better full-context prediction, diverging from the recency-dominated processing that humans actually do. Our multi-model comparison offers a complementary perspective: while Oh and Schuler varied training data size, we varied model capacity within the same training data and found that the effect remains detectable across model sizes.

The kinds of words that populate the off-diagonal cells (reported in the Results) point to what the two measures are tracking. The low-surprisal / high-extrapolation cell is enriched for coordinators and complementizers (e.g., ``and'', ``as'', ``that'', ``had''), which are lexically routine connectives that open a new syntactic constituent whose hidden-state trajectory will be different in kind from the preceding one. The high-surprisal / low-extrapolation cell contains both rarer content words that fit the established thematic frame (e.g., ``ocean'', ``manor'', ``tics'') and discourse pivots (``then'', ``however'', ``now'', ``first'') that are lexically unexpected at that position but whose structural role the trajectory has already begun to accommodate. The pattern suggests that the trajectory dimension is sensitive to structural continuity (frame, theme, syntactic commitment), whereas surprisal indexes lexical identifiability over the full context. These appear to be different psycholinguistic constructs, and the dissociation observed in our regressions is consistent with that distinction.

\subsection{Prediction, Trajectory, and the Nature of Comprehension}

An intriguing question raised by these findings concerns the causal relationship between prediction and trajectory. In language models, trajectory structure is a byproduct of optimizing next-token prediction. The model learns to predict words, and trajectory emerges in the hidden states because human-produced text has momentum. Surprisal is the primary quantity; trajectory is secondary.

Whether the same priority holds in human comprehenders is an open question. One possibility is that prediction and trajectory sensitivity are genuinely independent components of processing, each contributing its own cost. Another, more speculative possibility is that the causal direction differs in humans: that comprehension is fundamentally a dynamical process in which the evolving representational state carries local trajectory continuity that is actively maintained and exploited, and that sensitivity to word-level probability arises partly as a consequence of this process rather than as a fully independent computation. Words that are improbable tend to be trajectory-breaking, which is why surprisal predicts processing costs; but some portion of what surprisal captures may reflect trajectory disruption rather than prediction error per se. The present data are consistent with both interpretations, and with intermediate accounts in which prediction and trajectory interact in ways that neither tradition has fully specified.

What the data do establish is that the two measures are dissociable (nearly orthogonal in Natural Stories, with independent contributions to reading times) and that this dissociation calls for explanation. A direct experimental test, constructing stimuli matched on surprisal but varying in trajectory disruption or vice versa, would provide sharper evidence about the relationship. The theoretical question of how prediction and trajectory relate in human processing remains open and experimentally tractable.

\subsection{Limitations and Future Directions}

Several limitations should be noted. First, our reading-time data come from self-paced reading; eye-tracking data would provide a more sensitive test, particularly in distinguishing first-pass from reanalysis effects. Second, the garden-path analysis uses 24 items, and the contribution of trajectory extrapolation error is further attenuated in a crossed random-effects model that includes by-item intercepts. Because trajectory extrapolation error varies primarily between rather than within items at the disambiguation point, by-item intercepts absorb much of the variance the measure would otherwise explain. This is precisely the variance the theory predicts trajectory disruption should capture (sentences with larger trajectory disruptions should be harder), but with 24 items, the evidence cannot cleanly separate this account from item-level confounds. The Natural Stories replication addresses this limitation directly: with thousands of unique word positions across 10 diverse stories and 180 participants, the trajectory effect survives a fully crossed random-effects structure and actually strengthens when between-story variance is controlled, demonstrating that the effect is not an artifact of a few quirky garden-path items but a general property of incremental processing across naturalistic text.

The most promising future directions test the theoretical claims more directly. Constructing stimuli matched on surprisal but varying in trajectory disruption would provide the cleanest test of whether trajectory sensitivity is independent of word-level prediction. Neural measures (EEG, MEG) could establish whether trajectory extrapolation error predicts brain responses independently of surprisal, moving the evidence from behavior to mechanism.

A further implication concerns language generation. If local trajectory continuity contributes to processing ease, then human-produced and machine-produced text may differ in their trajectory statistics, and models or decoding strategies that preserve local trajectory coherence may yield text that is easier for humans to process. This remains speculative, but it offers a testable extension of the present framework.

\bibliographystyle{apalike}
\bibliography{references}

@inproceedings{biderman2023,
  author    = {Biderman, Stella and Schoelkopf, Hailey and Anthony, Quentin and Bradley, Herbie and O'Brien, Kyle and Hallahan, Eric and Khan, Mohammad Aflah and Purohit, Shivanshu and Prashanth, USVSN Sai and Raff, Edward and Skowron, Aviya and Sutawika, Lintang and van der Wal, Oskar},
  title     = {Pythia: A suite for analyzing large language models across training and scaling},
  booktitle = {Proceedings of the 40th International Conference on Machine Learning (ICML 2023)},
  year      = {2023}
}

@article{cho2017,
  author  = {Cho, Pyeong Whan and Goldrick, Matthew and Smolensky, Paul},
  title   = {Incremental parsing in a continuous dynamical system: Sentence processing in {G}radient {S}ymbolic {C}omputation},
  journal = {Linguistics Vanguard},
  volume  = {3},
  number  = {1},
  pages   = {20160105},
  year    = {2017}
}

@article{demberg2008,
  author  = {Demberg, Vera and Keller, Frank},
  title   = {Data from eye-tracking corpora as evidence for theories of syntactic processing complexity},
  journal = {Cognition},
  volume  = {109},
  number  = {2},
  pages   = {193--210},
  year    = {2008}
}

@article{ferreira2002,
  author  = {Ferreira, Fernanda and Swets, Benjamin},
  title   = {How incremental is language production? {E}vidence from the production of utterances requiring the computation of arithmetic sums},
  journal = {Journal of Memory and Language},
  volume  = {46},
  number  = {1},
  pages   = {57--84},
  year    = {2002}
}

@article{frank2015,
  author  = {Frank, Stefan L. and Otten, Leun J. and Galli, Giulia and Vigliocco, Gabriella},
  title   = {The {ERP} response to the amount of information conveyed by words in sentences},
  journal = {Brain and Language},
  volume  = {140},
  pages   = {1--11},
  year    = {2015}
}

@article{futrell2020,
  author  = {Futrell, Richard and Gibson, Edward and Levy, Roger P.},
  title   = {Lossy-context surprisal: An information-theoretic model of memory effects in sentence processing},
  journal = {Cognitive Science},
  volume  = {44},
  number  = {3},
  pages   = {e12814},
  year    = {2020}
}

@inproceedings{futrell2018,
  author    = {Futrell, Richard and Gibson, Edward and Tily, Harry J. and Blank, Idan and Vishnevetsky, Anastasia and Piantadosi, Steven T. and Fedorenko, Evelina},
  title     = {The {N}atural {S}tories corpus},
  booktitle = {Proceedings of LREC 2018},
  year      = {2018}
}

@article{gibson1998,
  author  = {Gibson, Edward},
  title   = {Linguistic complexity: Locality of syntactic dependencies},
  journal = {Cognition},
  volume  = {68},
  number  = {1},
  pages   = {1--76},
  year    = {1998}
}

@inproceedings{goodkind2018,
  author    = {Goodkind, Adam and Bicknell, Klinton},
  title     = {Predictive power of word surprisal for reading times is a linear function of language model quality},
  booktitle = {Proceedings of the 8th Workshop on Cognitive Modeling and Computational Linguistics},
  year      = {2018}
}

@inproceedings{hale2001,
  author    = {Hale, John},
  title     = {A probabilistic {E}arley parser as a psycholinguistic model},
  booktitle = {Proceedings of the Second Meeting of the North American Chapter of the Association for Computational Linguistics},
  year      = {2001}
}

@inproceedings{hewitt2019,
  author    = {Hewitt, John and Manning, Christopher D.},
  title     = {A structural probe for finding syntax in word representations},
  booktitle = {Proceedings of NAACL-HLT 2019},
  year      = {2019}
}

@article{huang2024,
  author  = {Huang, Yiding and Arehalli, Suhas and Kugemoto, Asa and Muxica, Michelle and Prasad, Pranali and Dillon, Brian and Linzen, Tal},
  title   = {Large-scale benchmark yields no evidence that language model surprisal explains syntactic disambiguation difficulty},
  journal = {Journal of Memory and Language},
  volume  = {137},
  pages   = {104510},
  year    = {2024}
}

@inproceedings{jawahar2019,
  author    = {Jawahar, Ganesh and Sagot, Beno\^{i}t and Seddah, Djam\'{e}},
  title     = {What does {BERT} learn about the structure of language?},
  booktitle = {Proceedings of ACL 2019},
  year      = {2019}
}

@book{levelt1989,
  author    = {Levelt, Willem J.M.},
  title     = {Speaking: From Intention to Articulation},
  publisher = {MIT Press},
  year      = {1989}
}

@article{levy2008,
  author  = {Levy, Roger},
  title   = {Expectation-based syntactic comprehension},
  journal = {Cognition},
  volume  = {106},
  number  = {3},
  pages   = {1126--1177},
  year    = {2008}
}

@article{lewis2005,
  author  = {Lewis, Richard L. and Vasishth, Shravan},
  title   = {An activation-based model of sentence processing as skilled memory retrieval},
  journal = {Cognitive Science},
  volume  = {29},
  number  = {3},
  pages   = {375--419},
  year    = {2005}
}

@inproceedings{oh2023a,
  author    = {Oh, Byung-Doh and Schuler, William},
  title     = {Transformer-based language model surprisal predicts human reading times best with about two billion training tokens},
  booktitle = {Findings of the Association for Computational Linguistics: EMNLP 2023},
  year      = {2023}
}

@article{oh2023b,
  author  = {Oh, Byung-Doh and Schuler, William},
  title   = {Why does surprisal from larger language models provide a poorer fit to human reading times?},
  journal = {Transactions of the Association for Computational Linguistics},
  volume  = {11},
  pages   = {336--350},
  year    = {2023}
}

@misc{radford2019,
  author = {Radford, Alec and Wu, Jeffrey and Child, Rewon and Luan, David and Amodei, Dario and Sutskever, Ilya},
  title  = {Language models are unsupervised multitask learners},
  howpublished = {OpenAI Blog},
  year   = {2019}
}

@inproceedings{roark2009,
  author    = {Roark, Brian and Bachrach, Asaf and Cardenas, Carlos and Pallier, Christophe},
  title     = {Deriving lexical and syntactic expectation-based measures for psycholinguistic modeling via incremental top-down parsing},
  booktitle = {Proceedings of the 2009 Conference on Empirical Methods in Natural Language Processing},
  pages     = {324--333},
  year      = {2009}
}

@article{schrimpf2021,
  author  = {Schrimpf, Martin and Blank, Idan A. and Tuckute, Greta and Kauf, Carina and Hosseini, Eghbal A. and Kanwisher, Nancy and Tenenbaum, Joshua B. and Fedorenko, Evelina},
  title   = {The neural architecture of language: Integrative modeling converges on predictive processing},
  journal = {Proceedings of the National Academy of Sciences},
  volume  = {118},
  number  = {45},
  pages   = {e2105646118},
  year    = {2021}
}

@article{smith2013,
  author  = {Smith, Nathaniel J. and Levy, Roger},
  title   = {The effect of word predictability on reading time is logarithmic},
  journal = {Cognition},
  volume  = {128},
  number  = {3},
  pages   = {302--319},
  year    = {2013}
}

@book{spivey2007,
  author    = {Spivey, Michael J.},
  title     = {The Continuity of Mind},
  publisher = {Oxford University Press},
  year      = {2007}
}

@article{tabor1999,
  author  = {Tabor, Whitney and Tanenhaus, Michael K.},
  title   = {Dynamical models of sentence processing},
  journal = {Cognitive Science},
  volume  = {23},
  number  = {4},
  pages   = {491--515},
  year    = {1999}
}

@inproceedings{wilcox2020,
  author    = {Wilcox, Ethan G. and Gauthier, Jon and Hu, Jennifer and Qian, Peng and Levy, Roger},
  title     = {On the predictive power of neural language models for human real-time comprehension behavior},
  booktitle = {Proceedings of the 42nd Annual Conference of the Cognitive Science Society},
  year      = {2020}
}

\end{document}